\documentclass[runningheads]{llncs}
\usepackage{graphicx}

\usepackage{tikz}
\usepackage{comment}
\usepackage{amsmath,amssymb} 
\usepackage{color}
\usepackage{multirow,array,makecell}
\usepackage{booktabs}
\usepackage[pagebackref=true,breaklinks=true,colorlinks,bookmarks=false]{hyperref}
\usepackage{url}
\usepackage{subcaption}
\usepackage[ruled]{algorithm2e}
\usepackage[htt]{hyphenat}  
\usepackage{graphbox}

\usepackage{cite}

\usepackage[accsupp]{axessibility}  

\newcommand\data[1]{{\normalfont \texttt{#1}}}
\newcommand\eg{\textit{e.g.}}
\newcommand\ie{\textit{i.e.}}
\usepackage{xspace}
\def\onedot{.\xspace}
\def\etal{\emph{et al}\onedot}

\def\fullname{Mutual Information Regularization with Oracle}
\def\method{MIRO}
\def\pacs{\data{PACS}}
\def\vlcs{\data{VLCS}}
\def\oh{\data{OfficeHome}}
\def\tr{\data{TerraIncognita}}
\def\dn{\data{DomainNet}}
\definecolor{darkred}{rgb}{0.8, 0.0, 0.0}

\DeclareMathOperator*{\argmin}{arg\,min}

\begin{document}
\pagestyle{headings}
\mainmatter
\def\ECCVSubNumber{3660}  

\title{Domain Generalization by Mutual-Information Regularization with Pre-trained Models}

\titlerunning{Domain Generalization by MI Regularization with Pre-trained Models}
%
\author{
Junbum Cha$^1$ \and
Kyungjae Lee$^2$ \and
Sungrae Park$^3$ \and
Sanghyuk Chun$^4$
}
\authorrunning{Cha et al.}
%
\institute{
$^1$ Kakao Brain \quad $^2$ Chung-Ang University \\
$^3$ Upstage AI Research \quad $^4$ NAVER AI Lab\\
\email{junbum.cha@kakaobrain.com, kyungjae.lee@ai.cau.ac.kr, \\ sungrae.park@upstage.ai, sanghyuk.c@navercorp.com}
}
\maketitle

\begin{abstract}
Domain generalization (DG) aims to learn a generalized model to an unseen target domain using only limited source domains. Previous attempts to DG fail to learn domain-invariant representations only from the source domains due to the significant domain shifts between training and test domains. Instead, we re-formulate the DG objective using mutual information with the oracle model, a model generalized to any possible domain. We derive a tractable variational lower bound via approximating the oracle model by a pre-trained model, called \fullname{} (\method{}). Our extensive experiments show that \method{} significantly improves the out-of-distribution performance. Furthermore, our scaling experiments show that the larger the scale of the pre-trained model, the greater the performance improvement of \method{}. 
Code is available at \url{https://github.com/kakaobrain/miro}.


\end{abstract}

\section{Introduction}
Emerging studies on the generalizability of deep neural networks have revealed that the existing models, which assume independent and identically distributed (i.i.d.) training and test distribution, are not robust to significant distribution shifts between training and test distribution, \eg, backgrounds \cite{xiao2020background_challenge_bgc}, geographic distribution \cite{de2019does}, demographic statistics \cite{yang2020towards, scimeca2022wcst-ml}, textures \cite{geirhos2019cnn_biased_towards_texture, bahng2019rebias}, or day-to-night shifts \cite{dai2018dark, michaelis2019benchmarking}.
Domain generalization (DG) aims to learn robust representations against distribution shifts from multiple source {domains} during training. The trained model is evaluated on an unseen domain to measure the robustness.
The existing DG approaches have tried to learn invariant features across multiple {domains} \cite{ganin2016dann,arjovsky2019irm,li2018mldg,cha2021swad,sun2016coral,bui2021mdsdi,zhou2021mixstyle}.
However, recent studies \cite{gulrajani2020domainbed, koh2021wilds} have shown that simple baselines without learning invariant features are comparable to or even outperform the existing DG methods on the diverse DG benchmarks with a fair evaluation protocol in realistic settings (\eg, using ResNet-50 instead of ResNet-18 \cite{he2016_cvpr_resnet}).
We presume that it is because training and test distributions differ too significantly to learn domain-invariant features by the training distribution only.

Instead of learning domain-invariant features, we let a model learn similar features to ``oracle'' representations, \ie, an optimal model generalized to \emph{any} domain. In particular, we re-formulate the DG problem by maximizing the mutual information (MI) between the oracle model representations and the target model representations while preserving the training loss on source domains. However, the oracle model is not achievable in practice. Hence, we use a large pre-trained model (\eg, ImageNet pre-trained ResNet-50 \cite{he2016_cvpr_resnet}) as an approximation. With this approximation, we derive a tractable variational lower bound of the proposed maximization problem, named \fullname{} (\method{}). At a high level, our \method{} objective consists of two objectives: an original target task (\ie, an ERM objective) and a regularization term between the pre-trained model and the current target model. Note that the standard DomainBed benchmark \cite{gulrajani2020domainbed} uses the ImageNet pre-trained ResNet-50 as the initialization of a DG method, thus, we use the pre-trained ResNet as the initialization and the approximation of the oracle model at the same time.

While a naive fine-tuning approach of a large pre-trained model can harm the robustness against distribution shifts \cite{wortsman2021robust, kumar2022iclr_finetuning_distort}, our proposed algorithm remarkably improves the robustness against unseen domains during fine-tuning in a plug-and-play manner to any scale of the backbone model and datasets. 
In our experiment, we observe that the naive fine-tuning of a larger pre-trained model can fail to provide better performances, even though the larger pre-trained model is trained with more data and domains. For example, ERM with the ResNet pre-trained on ImageNet (trained with 1.3M images) shows 64.2\% of averaged accuracy, while ERM with the ViT pre-trained on CLIP (trained with 400M image-caption pairs) shows 61.1\%.
On the other hand, we show that our method can significantly improve the average DG performances with backbone models at different scales, \eg, ImageNet pre-trained ResNet (64.2\% $\rightarrow$ 65.9\%), 400M image-text pre-trained ViT (CLIP) \cite{radford2021clip} (61.1\% $\rightarrow$ 73.7\%) and Instagram 3.6B pre-trained RegNet (SWAG) \cite{singh2022swag} (68.0\% $\rightarrow$ 74.1\%). Especially, we observe that the pre-trained knowledge by larger pre-trained models, such as SWAG and CLIP, is more effective to learn domain generalized features than the ImageNet pre-trained model: \method{} with the ViT pre-trained on CLIP outperforms \method{} with the ResNet pre-trained on ImageNet in contrast to the naive fine-tuning.
Furthermore, our feature-level regularization method is easily combined with the existing parameter space ensemble methods \cite{cha2021swad, wortsman2021robust} (74.1\% $\rightarrow$ \textbf{77.3\%} average DG accuracy by combining with SWAD \cite{cha2021swad} and pre-trained RegNet).

Our contribution is as follows: (1) We re-formulate the DG objective by mutual information with the oracle model. Then, we approximate the oracle by a large pre-trained model to derive a tractable approximation of the target objective. We propose \fullname{} (\method{}) to solve our objective. (2) We analyze the pre-trained models in terms of the MI with the oracle model. Our analysis shows that naive fine-tuning of pre-trained models can harm the MI with the oracle, on the other hand, \method{} shows high MI with the oracle.
(3) We compare \method{} with state-of-the-art DG methods on DomainBed. \method{} outperforms all methods in all settings, including varying optimizers and pre-trained models. We also provide extensive analysis to understand \method{}. For example, we observe that \method{} shows stronger DG performances with larger pre-trained models, such as SWAG \cite{singh2022swag} or CLIP \cite{radford2021clip}.


\section{Related works}

\subsubsection{Domain generalization.}
Learning domain-invariant features from source domains has been a major branch in the DG field. The main idea is discarding biased knowledge to a specific domain while preserving invariant features over source domains, by minimizing feature divergences between the source domains \cite{muandet2013icml_DIFL,ganin2016dann,li2018mmd,sun2016coral,matsuura2020aaai_mixture_mld,li2018cdann,zhao2020er_entropy_regularization}, simulating domain shifts based on meta-learning \cite{li2018mldg,balaji2018metareg,dou2019masf,li2019episodic_epi-fcr,zhang2020arm,bui2021mdsdi}, robust optimization \cite{arjovsky2019irm,krueger2020vrex,Sagawa2020GroupDRO, shi2022gradient,cha2021swad}, or augmenting source domain examples \cite{nuriel2020padain,zhou2021mixstyle,nam2019sagnet,carlucci2019jigsaw_jigen,zhou2020l2a_ot,bai2020decaug,robey2021mbdg_model_based_domain_generalization,yang2021atsl_adversarial_teacher_student_learning}.
However, even if the model learns invariant representation to source domains, it can still be biased toward the source domains which causes limited performance on unseen target domains. That is, learning invariant representation across source domains is not enough to achieve the underlying objective of domain generalization \cite{bui2021mdsdi,chattopadhyay2020dmg,chen2022preserving}. 
To compensate for the issue, this paper employs pre-trained models, which provide general representations across various domains including unseen target domains.



\subsubsection{Exploiting pre-trained models.}
There have been numerous attempts to exploit pre-trained models in various fields.
Transfer learning \cite{xuhong2018l2sp,li2019delta} and knowledge distillation \cite{ahn2019vid,tian2019crd} employ pre-trained models to improve in-domain performance when dataset or architecture shift occurs between pre-training and fine-tuning.
Continual learning utilizes the pre-trained model to maintain old task performance when learning new tasks \cite{li2017lwf}.
Recently, several studies targeting the out-of-distribution generalization are emerging \cite{kumar2022iclr_finetuning_distort,wortsman2021robust}. 
Kumar \etal \cite{kumar2022iclr_finetuning_distort} show that naive fine-tuning distorts the pre-trained features and propose a simple baseline, named LP-FT, to alleviate the distortion. 
WiSE-FT \cite{wortsman2021robust} focuses on zero-shot models. It combines pre-trained and fine-tuned weights to preserve the generalizability of the pre-trained zero-shot models.
In this paper, we propose a MI-based regularization method, \method{}, to exploit the generalizability of the pre-trained representation in the training process.

\section{Methods}

In this section, we first re-formulate the objective for the out-of-domain generalization by introducing an oracle model. Then, we derive a tractable variational bound of the objective by approximating the oracle model to the pre-trained model. The final form consists of the empirical risk and the mutual information (MI) regularization by querying the approximated oracle, named \fullname{} (\method{}). We empirically validate our approximation by MI between the oracle model and large pre-trained models. 

\subsection{Mutual information regularization with oracle}

The main idea of the proposed method is to guide the learning process using oracle representations of training datasets.
In general, the problem of domain generalization (DG) is to find a model that minimizes an expected loss of \emph{any} domain by using training datasets from only partial domains, which are called source domains.
Many existing methods minimize an empirical loss averaged over source domains.
More specifically, suppose that training samples $\{\mathcal{S}_{d}\}_{d=1}^{m}$ are given in $m$ domains and we consider a hypothesis set $\mathcal{H}$ for optimization.
Then, many existing DG frameworks can be formulated as follows:
\begin{align}\label{eq:origin_dg}
  \bar{h}=\argmin_{h\in\mathcal{H}} \sum_{d=1}^{m} \mathcal E_{\mathcal{S}_{d}}(h),
\end{align}
where $d$ indicates an individual source domain and $\mathcal E_{\mathcal{S}_{d}}$ is an empirical loss over the source domain $d$.
Note that majority of existing DG methods can be interpreted as the variant of Equation \eqref{eq:origin_dg}.
For example, if we choose a simple cross-entropy loss for $\mathcal E_{\mathcal{S}_{d}}$, then Equation \eqref{eq:origin_dg} becomes ``ERM'' baseline used in \cite{gulrajani2020domainbed}\footnote{Note that the terminology ERM can be unfair because other methods also minimize ``empirical risk'' but with different loss designs. We use the terminology ``ERM'' to indicate the cross-entropy baseline as suggested by Gulrajani and Lopez-Paz \cite{gulrajani2020domainbed}.}. Otherwise, $\mathcal E_{\mathcal{S}_{d}}$ can be formulated as a regularized ERM, such as IRM \cite{arjovsky2019irm} or CORAL \cite{sun2016coral}.
However, the formulation \eqref{eq:origin_dg} still suffers from learning domain-invariant representations using only partial domains when the target distribution differs significantly from the training distribution. For example, CORAL, the state-of-the-art method, shows inconsistent out-of-domain accuracies across domains in DomainNet \cite{peng2019domainnet}. While CORAL achieves $\approx$50\% top-1 accuracy on four \emph{easy} domains (59.2\% for Clipart, 46.6\% for Painting, 59.8\% for Real, 50.1\% for Sketches), it only shows 13.4\% for QuickDraw and 19.7\% for Infographics where the domains show the significant distribution shift comparing to others.

To alleviate this issue, we re-formulate the DG problem by employing \emph{oracle} representations of source domains. Here, we define an oracle model as a model that can be generalized to \emph{any} possible domain, not only for the source domains.
We define a model as a composition of a feature extractor $f$ and a classifier $g$ on the feature space where the whole classifier $h$ can be written as $h=f\circ g$.
Then, let $f^{*}$ be a feature extractor of the oracle model.
We first start from a strong assumption: 
we may assume that $f^*$ is accessible during the training phase. Then, we can obtain additional information from $f^{*}$ by querying the oracle representations of training samples in the source domains.
By using the oracle representations, we can guide the learning process of a target model by maximizing MI between oracle representations and target ones.
We formulate the proposed oracle-guided DG framework as follows:
\begin{align}
\begin{split} 
    \max_{h} \quad & I(Z_{f^\mathfrak{*}};Z_{f}) \\
    \textrm{s.t.} \quad & \mathcal{E}_{\mathcal S}(h) - \mathcal{E}_{\mathcal S}(\bar{h}) \leq \epsilon,
\end{split}
\label{eq:obj_constrained_mother}
\end{align}
where $Z_{f^{*}}$ is a random feature extracted by $f^{*}$ and $Z_{f}$ is a random feature extracted by a target model $f$. $I(Z_{f^{*}};Z_{f})$ is MI between $Z_{f^{*}}$ and $Z_{f}$, and $\mathcal{E}_{\mathcal S}(\cdot) = \sum_{d=1}^{m}\mathcal{E}_{\mathcal{S}_{d}}(\cdot)$.
The inequality constraint ensures the performance of the target model on the source domains.
Maximizing the MI will inhibit the target model from overfitting domain-specific features in the limited source domains. Because we assume that the ``oracle'' is generalized well to \textit{any} possible domain, the MI constraints \eqref{eq:obj_constrained_mother} will be beneficial to learning robust representations.


Unfortunately, the oracle feature extractor $f^{*}$ is not accessible in practice. Instead, we approximate the oracle feature extractor by using a pre-trained model $f^{0}$.
Our assumption is that a model pre-trained on large-scale diverse datasets, such as ImageNet \cite{russakovsky2015imagenet}, contains information on diverse domains.
In practice, we choose $f^{0}$ as the ImageNet pre-trained ResNet-50 \cite{he2016_cvpr_resnet}, the standard initialization choice for evaluating DG algorithms \cite{gulrajani2020domainbed}. We also consider models trained by larger diverse datasets, such as CLIP \cite{radford2021clip} (trained with 400M web crawled image-text pairs) and SWAG \cite{singh2022swag} (trained with 3.6B noisy image-hashtag pairs crawled from Instagram).
Although using CLIP and SWAG is not a fair comparison to the existing DG benchmark, here, we emphasize that naive fine-tuning of large pre-trained models leads to inferior generalizability to extreme distribution shifts at test time \cite{wortsman2021robust, kumar2022iclr_finetuning_distort}.
In our experiments, we also observe a similar observation: naive fine-tuning of CLIP shows an inferior DG performance (61.1\%) than ERM (64.2\%).

Through the approximation of the oracle model, we derive a tractable variational bound of our objective \eqref{eq:obj_constrained_mother}.
We assume a pre-trained model $f^{0}$ is located near $f^{*}$ in terms of distance equipped on the hypothesis set of the feature extractors and it can provide approximated representation of $f^{*}$.
Under this assumption, 
we can obtain a tractable objective by deriving an approximated lower bound of the MI.
We first derive the variational lower bound of the MI as follows:
\begin{align}
    I(Z_{f^{*}} ; Z_f)
    =&\mathbb{E}_{Z_{f^{*}}, Z_f}\left[\log \frac{q(Z_{f^{*}} \mid Z_f)}{p(Z_{f^{*}})}\right]+K L(p(Z_{f^{*}}\mid Z_f) \| q(Z_{f^{*}}\mid Z_f)) \nonumber\\
    \geq& \mathbb{E}_{Z_{f^{*}}, Z_f}[\log q(Z_{f^{*}}\mid Z_f)]+H(Z_{f^{*}}),\label{eq:variational_lower_bnd}
\end{align}
where $q$ is the variational distribution with a mild regularity condition.
More detailed derivation can be found in Barber and Agakov \cite{agakov2004vim}.
Then, we approximate the expectation in Equation \eqref{eq:variational_lower_bnd} by using $f^{0}$. 
\begin{align}
    I(Z_{f^{*}} ; Z_f) &\geq \mathbb{E}_{Z_{f^{*}}, Z_f}\left[\log q(Z_{f^{*}} \mid Z_f)\right]+H(Z_{f^{*}})\nonumber\\
    &\geq \mathbb{E}_{Z_{f^{0}},Z_{f}}\left[\log q(Z_{f^{0}} \mid Z_{f})\right] - Cd_{2,\infty}(f^{*}, f^{0})+H(Z_{f^{*}}),\label{eq:distribution_change}
\end{align}
where $C$ is a constant and $d_{2,\infty}(f^{*},f^{0}):=\sup_{x}\|f^{*}(x)-f^{0}(x)\|_{2}$.
Note that $d_{2,\infty}$ is a proper metric on the hypothesis set of feature extractor.
The last inequality of Equation \eqref{eq:distribution_change} is derived by using the first-order Taylor expansion and assuming the regularity condition of $q$ (See Appendix \ref{sec:derivation_of_lower_bound}).
We would like to note that the inequality is tight enough due to Taylor's theorem.
In other words, equality condition of the last inequality of Equation \eqref{eq:distribution_change} is $d_{2,\infty}(f^{*}, f^{0})=0$.
Hence, $d_{2,\infty}(f^{*}, f^{0})$ represents the effect of the pre-trained model $f^{0}$ on the approximation of the lower bound.
Intuitively speaking, the lower bound shows that the smaller $d_{2,\infty}(f^{*}, f^{0})$ is, the tighter the gap between the true lower bound and approximated one is.
In summary, the MI between $Z_{f^{*}}$ and $Z_{f}$ can be maximized by maximizing the term $\mathbb{E}_{Z_{f^{0}}, Z_f}[\log q(Z_{f^{0}}\mid Z_f)]$.

Finally, to consider the constraint term, we introduce the Lagrangian method to Equation \eqref{eq:obj_constrained_mother}, then we can derive an objective function from Equation  \eqref{eq:distribution_change}:
\begin{align}
    R(h)=\mathbb{E}_{Z_{f^0}, Z_f}[\log q(Z_{f^0} \mid Z_f)] - \beta \mathcal{E}_{\mathcal S}(h),
\end{align}
where $\beta$ indicates the Lagrangian multiplier.
Note that the entropy of $Z_{f^{*}}$ and $d_{2,\infty}(f^{*},f^{0})$ are omitted, since they are independent to our optimization target $h=f\circ g$.
In the implementation, we model the variational distribution as a Gaussian distribution with mean vector $\mu(Z_{f})$ and covariance matrix $\Sigma(Z_{f})$ and replace the multiplier $\beta$ with the regularization coefficient $\lambda$.
Then, our final loss function becomes:
\begin{align}
    \textbf{(MIRO)} \quad \mathcal L(h) = \mathcal{E}_{\mathcal S}(h) + \lambda \mathbb{E}_{Z_{f^{0}},Z_{f}}\left[ \log \left|\Sigma(Z_{f})\right| + \|Z_{f^{0}}-\mu(Z_{f})\|^{2}_{\Sigma(Z_{f})^{-1}}\right],
    \label{eq:method}
\end{align}
where $\|x\|_{A}=\sqrt{x^{\intercal}A x}$ and constants independent on $h$ are omitted.
Then, we optimize the loss function using  a stochastic gradient method.
The entire learning process is summarized in Algorithm \ref{algorithm}.
In the following sections, we empirically justify our approximation of $f^{*}$ and explain implementation details for the mean and variance encoders of the Gaussian distribution $q$.

\SetKwInput{KwInit}{Init}
\begin{algorithm}[t!]
    \DontPrintSemicolon
    \caption{\fullname{} (\method{})}
    \label{algorithm}

    \KwIn{feature extractor $f$, classifier $g$, mean encoder $\mu$, variance encoder $\Sigma$, regularization coefficient $\lambda$, batch size $N$.}
    \KwInit{initialize $f$ to pre-trained feature extractor $f^0$.}
    \KwOut{learned feature extractor $f$ and learned classifier $g$.}
    
    \For{\textup{sampled mini-batch ($\mathbf{x, y}$)}}{
        $\mathbf z_f=f(\mathbf x)$ \\
        $\mathbf z_{f^0}=f^0(\mathbf x)$ \\
        $\mathcal L=\frac{1}{N}\sum_i^N \left[ \texttt{CrossEntropy}\left(g(z^i_f), y^i\right) + \lambda\left( \log \left|\Sigma(z_{f}^{i})\right| + \|z_{f^{0}}^{i}-\mu(z_{f}^{i})\|^{2}_{\Sigma(z_{f}^{i})^{-1}}\right) \right] $ \\
        \textup{update $f, g, \mu, \Sigma$ to minimize $\mathcal L$}
    }
\end{algorithm}

\begin{figure}
    \centering
    \begin{subfigure}{0.49\linewidth}
        \includegraphics[width=\linewidth]{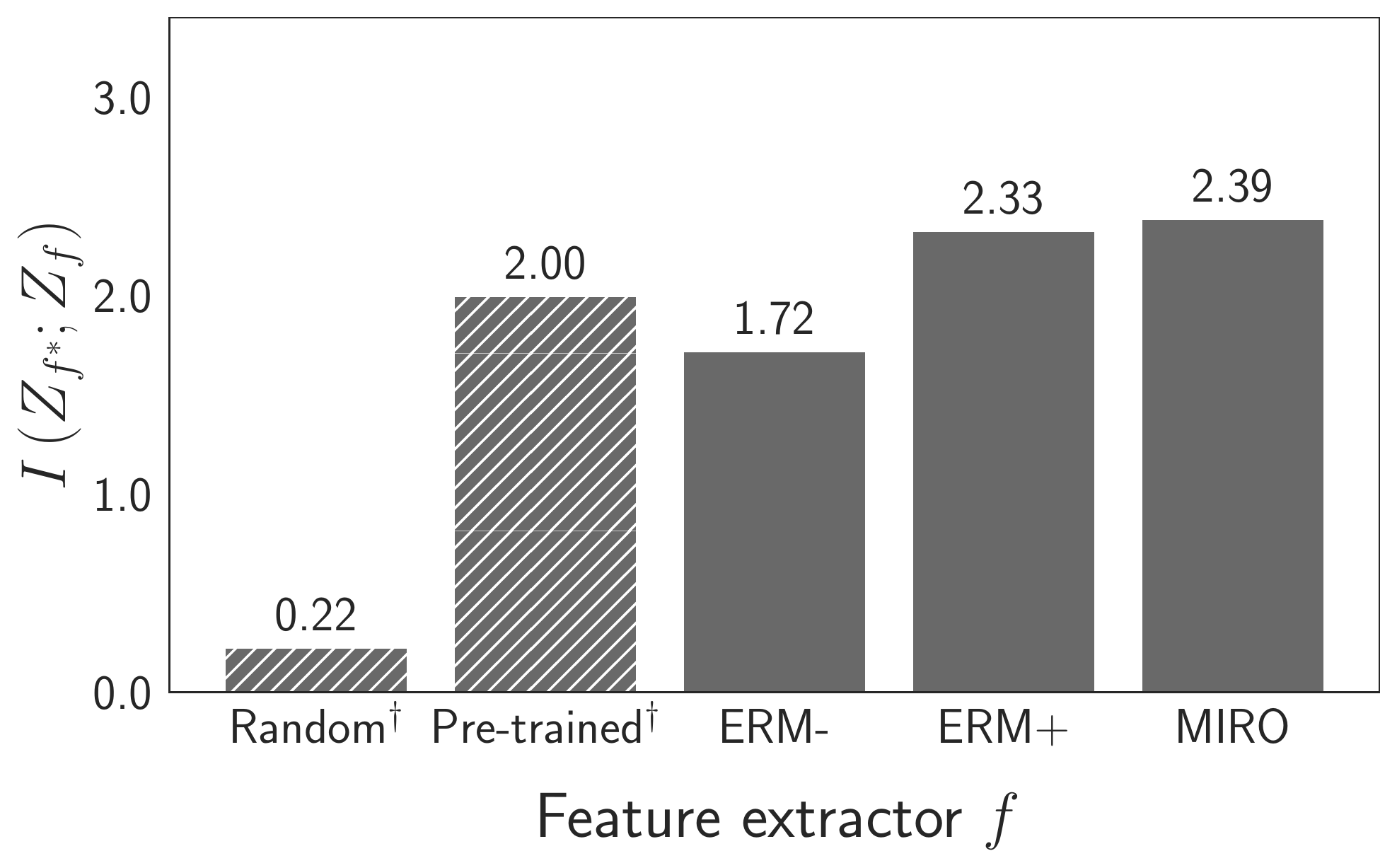}
        \caption{\small ResNet-50}
        \label{fig:mi-resnet}
    \end{subfigure}%
    \hfill
    \begin{subfigure}{0.49\linewidth}
        \includegraphics[width=\linewidth]{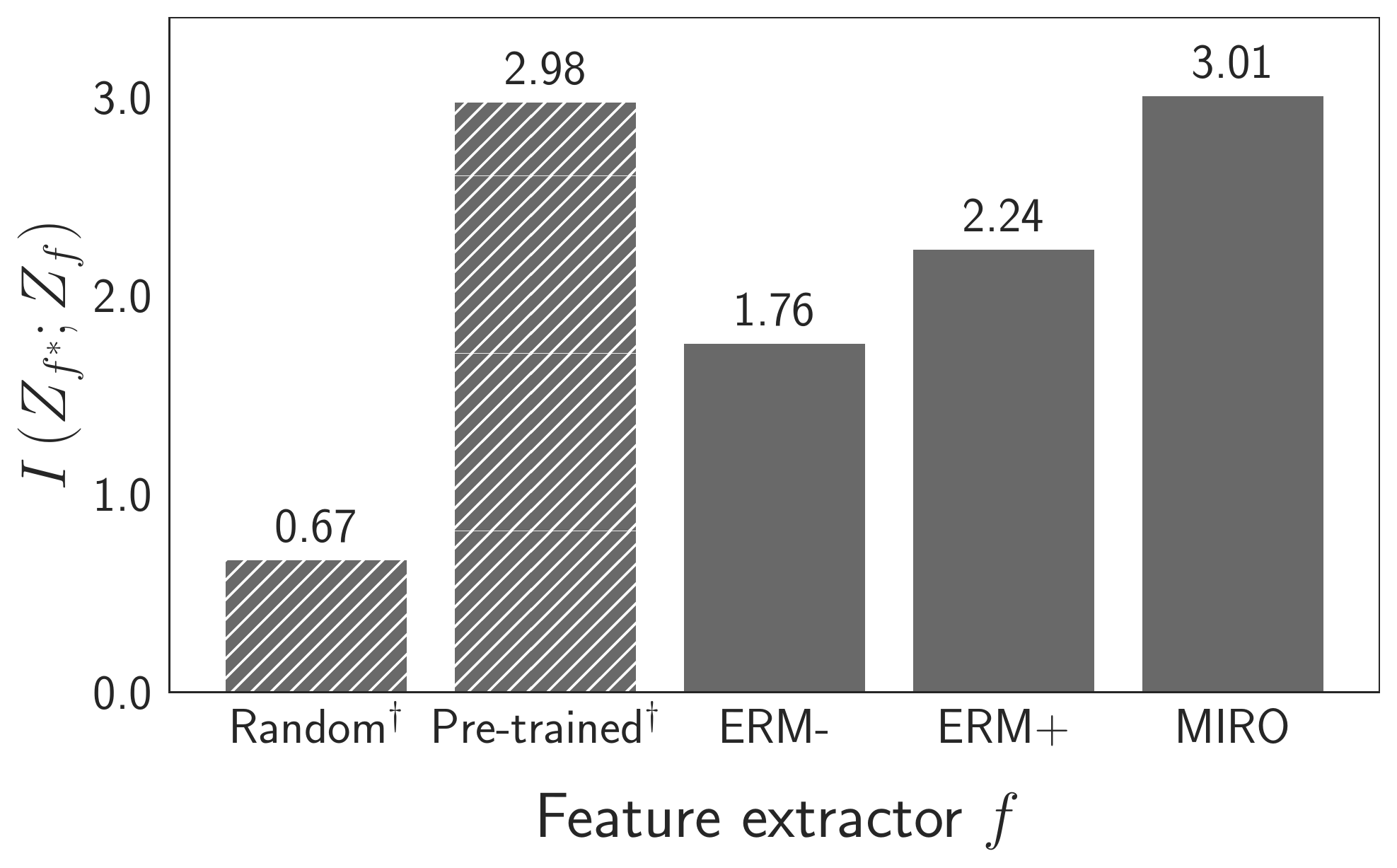}
        \caption{\small RegNetY-16GF}
        \label{fig:mi-regnet}
    \end{subfigure}%
    \caption{\small \textbf{Mutual information $I\left( Z_{f^{*}}; Z_f \right)$ with oracle model.} The mutual information is estimated by MINE~\cite{belghazi2018mine} in \data{PACS}. Oracle model is trained using all of the four domains. \textit{Random} and \textit{Pre-trained} indicate random and pre-trained model initialization, respectively. \textit{ERM-} and \textit{ERM+} are trained from random and pre-trained model initialization, respectively. $\dagger$ indicates models without fine-tuning. The experiments are repeated with two pre-trained models: ImageNet 1.3M pre-trained ResNet-50 and Instagram 3.6B pre-trained RegNetY-16GF.}
    \label{fig:mi}
\end{figure}


\subsection{Mutual information analysis with the oracle model}
\label{subsection:mi_with_oracle}

Here, we empirically show how our approximation by pre-trained models is close to the oracle model and how our algorithm is effective to learn representations having high mutual information (MI) to the underlying oracle model. More specifically, we compare MI between the candidate models and the oracle model on the \pacs{} dataset.
Since the \emph{true} oracle model 
is not achievable in practice, we train an oracle model by directly optimizing a model on the entire domains. We train two oracle models with ResNet-50 and RegNetY-16GF backbones, where the average validation accuracies across all domains
are $97.2\%$ and $98.4\%$, respectively. We estimate MI between models by mutual information neural estimation (MINE) \cite{belghazi2018mine}. We describe the full details in Appendix \ref{subsec:mine}.

Figure \ref{fig:mi} illustrates the empirical MI between the candidate models and the oracle model.
In the figures, we first observe that the larger and more powerful pre-trained backbone (``Pre-trained'' in Figure \ref{fig:mi-regnet}) shows higher MI than the smaller backbone (``Pre-trained'' in Figure \ref{fig:mi-resnet}). Both pre-trained models consistently outperform ``Random'' in MI regardless of the backbone models. Our observations imply that a larger and stronger model is closer to the oracle model in terms of MI.
Similarly, we observe that ERM$+$ always shows high MI than ERM$-$. However, interestingly, in Figure \ref{fig:mi-regnet}, we observe that fine-tuning significantly harms MI of the pre-trained model (``Pre-trained'' vs. ``ERM$+$'') when the pre-trained model becomes larger and more powerful. Our observation is aligned in the same line as the previous studies on fine-tuning of large models \cite{wortsman2021robust, kumar2022iclr_finetuning_distort}.
Lastly, in both scenarios of ImageNet pre-trained ResNet (Figure \ref{fig:mi-resnet}) and SWAG pre-trained RegNet (Figure \ref{fig:mi-regnet}), our \method{} shows the highest MI with the oracle model.
Note that MI with the oracle model may not be completely aligned with the DG performance, but in practice, we observed that the evaluation ranking of the candidates is the same as the MI ranking; \method{} scores the best, followed by ERM$+$ and ERM$-$. Detailed results are provided in Appendix \ref{subsec:relationship_mi_dg}.

\subsection{Features and encoders design}

\subsubsection{Multi-scale features.}
One can only use the last-level features for our regularization.
However, high-level features can include pre-training task-related information, often irrelevant to the target task. 
Instead,
we use the intermediate outputs by each model block, \ie, stem output, blocks 1, 2, 3, and 4 for ResNet \cite{he2016_cvpr_resnet} and RegNet \cite{radosavovic2020regnet}, and stem output, blocks 3, 6, 9, and 12 for ViT-B.

\subsubsection{Design of the mean and variance encoders.} The multi-level structure increases the feature size, resulting in a computational cost increase. 
We alleviate the issue by employing 
simple yet effective architectures, identity function for the mean encoder and a bias-only model with diagonal covariance for the variance encoder.
We also tested more complicated architectures, but only computational cost was increased without performance improvement.

\section{Experiments}

\subsection{Experiment setups and implementation details}


\subsubsection{Evaluation protocols and datasets.} 
We employ DomainBed evaluation protocols \cite{gulrajani2020domainbed, cha2021swad} for a fair comparison. The five benchmark datasets are used: \data{PACS}~\cite{li2017pacs} (4 domains, 7 classes, and $9,991$ images), \data{VLCS}~\cite{fang2013vlcs} (4 domains, 5 classes, and $10,729$ images), \data{OfficeHome}~\cite{venkateswara2017officehome} (4 domains, 65 classes, and $15,588$ images), \data{TerraIncognita}~\cite{beery2018terraincognita} (4 domains, 10 classes, and $24,788$ images), and \data{DomainNet}~\cite{peng2019domainnet} (6 domains, 345 classes, and $586,575$ images).
All performance scores are evaluated by \textit{leave-one-out cross-validation}, where averaging all cases that use a single domain as the target (test) domain and the others as the source (training) domains. Every experiment is repeated three times. We leave 20\% of source domain data for validation. We use training-domain validation for the model selection and the hyperparameter search following DomainBed \cite{gulrajani2020domainbed}.

\subsubsection{Implementation details.}
We use ResNet-50 \cite{he2016_cvpr_resnet} pre-trained in the ImageNet \cite{russakovsky2015imagenet} as default. The model is optimized using Adam \cite{kingma2015adam} optimizer. A mini-batch contains all domains and 32 examples per domain.
The regularization coefficient $\lambda$ is tuned in [1.0, 0.1, 0.01, 0.001].
The other hyperparameters, such as batch size, learning rate, dropout rate, and weight decay, are tuned in the similar search space proposed in Cha \etal \cite{cha2021swad}. We provide full details in Appendix \ref{sec:additional_impl_details}.




\subsection{Main results}

\begin{table}
\centering
\small
\caption{\small {\bf Comparison with domain generalization methods.} Out-of-domain accuracies on five domain generalization benchmarks are shown. 
We highlight the \textbf{best results} in bold. 
The results marked by $\dagger, \ddagger$ are the reported numbers from Gulrajani and Lopez-Paz \cite{gulrajani2020domainbed} and Cha \etal \cite{cha2021swad}, respectively. The results of Fish, SelfReg, and mDSDI are the reported ones from each paper. Average accuracies and standard errors are reported from three trials.
}
\label{table:main_table}
\renewcommand{\arraystretch}{1.1}
\setlength{\tabcolsep}{3pt}
\setlength{\abovetopsep}{0.5em}
\begin{tabular}{l|ccccc|c}
\toprule
\textbf{Algorithm} & \data{PACS}          & \data{VLCS}          & \data{OfficeHome}    & \data{TerraInc}      & \data{DomainNet}     & {Avg.}  \\
\midrule
MMD$^\dagger$ \cite{li2018mmd}                  & 84.7\scriptsize{$\pm0.5$}          & 77.5\scriptsize{$\pm0.9$}          & 66.3\scriptsize{$\pm0.1$}          & 42.2\scriptsize{$\pm1.6$}          & 23.4\scriptsize{$\pm9.5$}          & 58.8           \\
Mixstyle$^\ddagger$ \cite{zhou2021mixstyle}     & 85.2\scriptsize{$\pm0.3$}          & 77.9\scriptsize{$\pm0.5$}          & 60.4\scriptsize{$\pm0.3$}          & 44.0\scriptsize{$\pm0.7$}          & 34.0\scriptsize{$\pm0.1$}          & 60.3           \\
GroupDRO$^\dagger$ \cite{Sagawa2020GroupDRO}    & 84.4\scriptsize{$\pm0.8$}          & 76.7\scriptsize{$\pm0.6$}          & 66.0\scriptsize{$\pm0.7$}          & 43.2\scriptsize{$\pm1.1$}          & 33.3\scriptsize{$\pm0.2$}          & 60.7           \\
IRM$^\dagger$ \cite{arjovsky2019irm}            & 83.5\scriptsize{$\pm0.8$}          & 78.5\scriptsize{$\pm0.5$}          & 64.3\scriptsize{$\pm2.2$}          & 47.6\scriptsize{$\pm0.8$}          & 33.9\scriptsize{$\pm2.8$}          & 61.6           \\
ARM$^\dagger$ \cite{zhang2020arm}               & 85.1\scriptsize{$\pm0.4$}          & 77.6\scriptsize{$\pm0.3$}          & 64.8\scriptsize{$\pm0.3$}          & 45.5\scriptsize{$\pm0.3$}          & 35.5\scriptsize{$\pm0.2$}          & 61.7           \\
VREx$^\dagger$ \cite{krueger2020vrex}           & 84.9\scriptsize{$\pm0.6$}          & 78.3\scriptsize{$\pm0.2$}          & 66.4\scriptsize{$\pm0.6$}          & 46.4\scriptsize{$\pm0.6$}          & 33.6\scriptsize{$\pm2.9$}          & 61.9           \\
CDANN$^\dagger$ \cite{li2018cdann}              & 82.6\scriptsize{$\pm0.9$}          & 77.5\scriptsize{$\pm0.1$}          & 65.8\scriptsize{$\pm1.3$}          & 45.8\scriptsize{$\pm1.6$}          & 38.3\scriptsize{$\pm0.3$}          & 62.0           \\
DANN$^\dagger$ \cite{ganin2016dann}             & 83.6\scriptsize{$\pm0.4$}          & 78.6\scriptsize{$\pm0.4$}          & 65.9\scriptsize{$\pm0.6$}          & 46.7\scriptsize{$\pm0.5$}          & 38.3\scriptsize{$\pm0.1$}          & 62.6           \\
RSC$^\dagger$ \cite{huang2020rsc}               & 85.2\scriptsize{$\pm0.9$}          & 77.1\scriptsize{$\pm0.5$}          & 65.5\scriptsize{$\pm0.9$}          & 46.6\scriptsize{$\pm1.0$}          & 38.9\scriptsize{$\pm0.5$}          & 62.7           \\
MTL$^\dagger$ \cite{blanchard2021mtl_marginal_transfer_learning}    & 84.6\scriptsize{$\pm0.5$}          & 77.2\scriptsize{$\pm0.4$}          & 66.4\scriptsize{$\pm0.5$}          & 45.6\scriptsize{$\pm1.2$}          & 40.6\scriptsize{$\pm0.1$}          & 62.9           \\
Mixup$^\dagger$ \cite{xu2020interdomain_mixup_aaai,yan2020interdomain_mixup,wang2020interdomain_mixup_icassp_dg}             & 84.6\scriptsize{$\pm0.6$}          & 77.4\scriptsize{$\pm0.6$}          & 68.1\scriptsize{$\pm0.3$}          & 47.9\scriptsize{$\pm0.8$}          & 39.2\scriptsize{$\pm0.1$}          & 63.4           \\
MLDG$^\dagger$ \cite{li2018mldg}                & 84.9\scriptsize{$\pm1.0$}          & 77.2\scriptsize{$\pm0.4$}          & 66.8\scriptsize{$\pm0.6$}          & 47.7\scriptsize{$\pm0.9$}          & 41.2\scriptsize{$\pm0.1$}          & 63.6           \\
Fish \cite{shi2022gradient}                     & 85.5\scriptsize{$\pm0.3$}          & 77.8\scriptsize{$\pm0.3$}          & 68.6\scriptsize{$\pm0.4$}          & 45.1\scriptsize{$\pm1.3$}          & 42.7\scriptsize{$\pm0.2$}          & 63.9           \\ 
ERM$^\ddagger$ \cite{vapnik1998statistical}     & 84.2\scriptsize{$\pm0.1$}          & 77.3\scriptsize{$\pm0.1$}          & 67.6\scriptsize{$\pm0.2$}          & 47.8\scriptsize{$\pm0.6$}          & 44.0\scriptsize{$\pm0.1$}          & 64.2           \\
SagNet$^\dagger$ \cite{nam2019sagnet}             & \textbf{86.3\scriptsize{$\pm0.2$}} & 77.8\scriptsize{$\pm0.5$}          & 68.1\scriptsize{$\pm0.1$}          & 48.6\scriptsize{$\pm1.0$}          & 40.3\scriptsize{$\pm0.1$}          & 64.2           \\
SelfReg \cite{kim2021selfreg}                   & 85.6\scriptsize{$\pm0.4$}          & 77.8\scriptsize{$\pm0.9$}          & 67.9\scriptsize{$\pm0.7$}          & 47.0\scriptsize{$\pm0.3$}          & 42.8\scriptsize{$\pm0.0$}          & 64.2           \\ 
CORAL$^\dagger$ \cite{sun2016coral}             & 86.2\scriptsize{$\pm0.3$}          & 78.8\scriptsize{$\pm0.6$}          & 68.7\scriptsize{$\pm0.3$}          & 47.6\scriptsize{$\pm1.0$}          & 41.5\scriptsize{$\pm0.1$}          & 64.5           \\
mDSDI \cite{bui2021mdsdi}              & 86.2\scriptsize{$\pm0.2$}          & \textbf{79.0\scriptsize{$\pm0.3$}}          & 69.2\scriptsize{$\pm0.4$}          & 48.1\scriptsize{$\pm1.4$}          & 42.8\scriptsize{$\pm0.1$}          & 65.1           \\ 
\textbf{\method}      & 85.4\scriptsize{$\pm0.4$}          & \textbf{79.0\scriptsize{$\pm0.0$}} & \textbf{70.5\scriptsize{$\pm0.4$}} & \textbf{50.4\scriptsize{$\pm1.1$}} & \textbf{44.3\scriptsize{$\pm0.2$}} & \textbf{65.9}  \\
\midrule
\multicolumn{7}{l}{\textit{Combined with SWAD \cite{cha2021swad}}}   \\
\midrule
ERM + SWAD$^\ddagger$                & 88.1\scriptsize{$\pm0.1$}          & 79.1\scriptsize{$\pm0.1$}          & 70.6\scriptsize{$\pm0.2$}          & 50.0\scriptsize{$\pm0.3$}          & 46.5\scriptsize{$\pm0.1$}          & 66.9           \\
CORAL + SWAD$^\ddagger$              & 88.3\scriptsize{$\pm0.1$}          & 78.9\scriptsize{$\pm0.1$}          & 71.3\scriptsize{$\pm0.1$}          & 51.0\scriptsize{$\pm0.1$}          & 46.8\scriptsize{$\pm0.0$}          & 67.3           \\
\textbf{\method{} + SWAD}            & \textbf{88.4\scriptsize{$\pm0.1$}} & \textbf{79.6\scriptsize{$\pm0.2$}} & \textbf{72.4\scriptsize{$\pm0.1$}} & \textbf{52.9\scriptsize{$\pm0.2$}} & \textbf{47.0\scriptsize{$\pm0.0$}} & \textbf{68.1} \\
\midrule
\multicolumn{7}{l}{\textit{Using RegNetY-16GF backbone with SWAG pre-training \cite{singh2022swag}}}   \\ \midrule
ERM & 89.6\scriptsize{$\pm0.4$}  &  78.6\scriptsize{$\pm0.3$}  &  71.9\scriptsize{$\pm0.6$}  &  51.4\scriptsize{$\pm1.8$}  &  48.5\scriptsize{$\pm0.6$} & 68.0 \\
\textbf{\method}        & \textbf{97.4\scriptsize{$\pm0.2$}}  &  \textbf{79.9\scriptsize{$\pm0.6$}}  &  \textbf{80.4\scriptsize{$\pm0.2$}}  &  \textbf{58.9\scriptsize{$\pm1.3$}}  &  \textbf{53.8\scriptsize{$\pm0.1$}}  &  \textbf{74.1}  \\
\midrule
ERM + SWAD      & 94.7\scriptsize{$\pm0.2$}  &  79.7\scriptsize{$\pm0.2$}  &  80.0\scriptsize{$\pm0.1$}  &  57.9\scriptsize{$\pm0.7$}  &  53.6\scriptsize{$\pm0.6$}  &  73.2  \\
\textbf{\method{} + SWAD}  & \textbf{96.8\scriptsize{$\pm0.2$}}  &  \textbf{81.7\scriptsize{$\pm0.1$}}  &  \textbf{83.3\scriptsize{$\pm0.1$}}  &  \textbf{64.3\scriptsize{$\pm0.3$}}  &  \textbf{60.7\scriptsize{$\pm0.0$}}  &  \textbf{77.3}  \\
\bottomrule
\end{tabular}
\end{table}
\subsubsection{Comparison with domain generalization methods.}
We provide exhaustive out-of-domain performance comparisons on five DG benchmarks in Table~\ref{table:main_table}. 
Compared to ERM, the proposed MI regularization significantly improves performance on every benchmark dataset, resulting in +1.7pp average improvement. 
Compared with the state-of-the-art methods, \method{} achieves the best performances in all benchmarks, except \pacs. 
Especially, \method{} remarkably outperforms previous methods: +1.3pp in \oh{} (mDSDI \cite{bui2021mdsdi}; $69.2\% \rightarrow 70.5\%$) and +1.8pp in \tr{} (SagNet \cite{nam2019sagnet}; $48.6\% \rightarrow 50.4\%$).
Considering the extensive experiment setup with 5 datasets and 22 target domains, the results demonstrate the effectiveness of \method{} to the diverse visual data types.

The second part of Table~\ref{table:main_table} shows the performance with stochastic weight averaging densely (SWAD)~\cite{cha2021swad}, a state-of-the-art optimizer for DG by seeking flat minima. Since SWAD is an orthogonal direction to \method{}, we also evaluate the combination of \method{} and SWAD.
As shown in the table, the combination of \method{} and SWAD achieves the best performance in all datasets, resulting in +0.8pp average improvement compared to the previous best results.

In the last part of Table \ref{table:main_table}, we push the limits of the out-of-domain performance by employing a large-scale backbone, RegNetY-16GF pre-trained by SWAG \cite{singh2022swag}; a weakly-supervised pre-trained model using 3.6 billion noisy Instagram images and hashtags. As shown in our previous study on MI with the oracle model, the pre-trained RegNet has higher MI than ImageNet pre-trained ResNet (Figure \ref{fig:mi}). In the experiments, we first observe that the improvement gap by \method{} becomes remarkably large compared to the ResNet pre-trained model (from +1.7pp to +6.1pp). We presume that this significantly large gap originated from the negative effect of naive fine-tuning as observed by previous works \cite{wortsman2021robust, kumar2022iclr_finetuning_distort} and our study (Figure \ref{fig:mi-regnet}). As shown in Figure \ref{fig:mi-regnet}, \method{} keeps MI with the oracle model high, resulting in remarkable performance gains on large-scale models. We further explore the effect of the scalability of pre-trained models in the later section.
Finally, by combining \method{} with RegNet backbone and SWAD, we achieve the best domain generalization results (77.3\%) on our evaluation benchmark.

\begin{table}[t]
\centering
\small
\caption{\small {\bf Comparison with various pre-training datasets, methods, and backbones.} We compare the performance changes according to the scale of the dataset, the method, and the backbone architecture of pre-training. ResNet-50 architecture is used as default. \data{OH}, \data{TI}, and \data{DN} indicate \data{OfficeHome}, \data{TerraIncognita}, and \data{DomainNet}, respectively. Every accuracy is averaged over three trials.
}
\label{table:varying_dataset_method_backbone_scale}
\renewcommand{\arraystretch}{1.1}
\setlength{\tabcolsep}{3pt}
\setlength{\abovetopsep}{0.5em}
\begin{tabular}{lll|ccccc|c} 
\toprule
\textbf{Dataset \scriptsize{(size)}} & \textbf{Pre-training}  & \textbf{Alg.} & \data{PACS} & \data{VLCS} & \data{OH} & \data{TI} & \data{DN} & {Avg.}  \\
\midrule\multirow{6}{*}{ImageNet \scriptsize{(1.3M)}}     & \multirow{2}{*}{ERM}   & ERM             & 84.2          & 77.3          & 67.6                & 47.8              & 44.0             & 64.2           \\
                                             &                        & \method         & 85.4          & 79.0          & 70.5                & 50.4              & 44.3             & 65.9 (+1.7)           \\
\cmidrule{2-9}& \multirow{2}{*}{Barlow Twins}                         & ERM             & 78.7          & 77.3          & 57.6                & 36.9              & 41.7             & 58.4           \\
                                    &                                 & \method         & 80.7          & 79.4          & 63.7                & 43.2              & 42.6             & 61.9 (+3.5)           \\
\cmidrule{2-9}& \multirow{2}{*}{MoCo v3}                              & ERM             & 86.7          & 77.3          & 61.8                & 49.1              & 43.8             & 63.7           \\
                                    &                                 & \method         & 86.3          & 78.5          & 66.8                & 48.4              & 44.7             & 65.0 (+1.3)           \\
\midrule\multirow{4}{*}{CLIP \scriptsize{(400M)}}        & \multirow{2}{*}{CLIP \scriptsize{(ResNet)}}   & ERM             & 64.3          & 69.8          & 28.2                & 32.9              & 29.5             & 44.9           \\
                                    &                                 & \method         & 76.6          & 78.9          & 59.5                & 49.0              & 42.0             & 61.2 (+16.3)           \\
\cmidrule{2-9}& \multirow{2}{*}{CLIP \scriptsize{(ViT)}}              & ERM             & 83.4          & 75.9          & 66.4                & 35.3              & 44.4             & 61.1           \\
                                    &                                 & \method         & 95.6          & 82.2          & 82.5                & 54.3              & 54.0             & 73.7 (+12.6)           \\
\midrule
\multirow{2}{*}{Instagram \scriptsize{(3.6B)}}  & \multirow{2}{*}{SWAG \scriptsize{(RegNet)}} & ERM      & 89.6          & 78.6          & 71.9                & 51.4              & 48.5               & 68.0           \\
                                   &                                             & \method  & 97.4          & 79.9          & 80.4                & 58.9              & 53.8               & 74.1 (+6.1)           \\
\bottomrule
\end{tabular}
\end{table}
\subsubsection{\method{} with various pre-trained models.} 
In this subsection, we investigate the robustness of the proposed method to the choice of pre-trained models.
In Table~\ref{table:varying_dataset_method_backbone_scale}, we explore the performance changes of \method{} by varying pre-training datasets, methods, and backbones.
From the pre-training method perspective, we examine two image self-supervised pre-training methods (Barlow Twins \cite{zbontar2021barlowtwins} and MoCo v3 \cite{chen2021mocov3}), one image-language self-supervised pre-training method (CLIP \cite{radford2021clip}), and one weakly-supervised pre-training method (SWAG \cite{singh2022swag}), as well as ImageNet supervised pre-training baseline (ImageNet ERM). 
From the pre-training scale perspective, we employ the ImageNet \cite{russakovsky2015imagenet} dataset of 1.3 million examples, the CLIP dataset of 400 million examples, and the Instagram dataset of 3.6 billion examples. 
We use ResNet-50 \cite{he2016_cvpr_resnet} backbone architecture as default, but a bigger model is also used for the large-scale pre-training, such as ViT-B \cite{dosovitskiy2021vit} for CLIP or RegNetY-16GF \cite{radosavovic2020regnet} for SWAG.

As shown in the table, \method{} improves performances compared with the baseline ERM in all experiments. For the ImageNet pre-training, applying \method{} results in performance improvements of +1.7pp, +3.5pp, and +1.3pp for ERM (supervised learning), Barlow Twins, and MoCo v3, respectively. For the large-scale pre-training, such as CLIP and SWAG, \method{} brings larger performance improvements of +16.3pp, +12.6pp, and +6.1pp for CLIP, CLIP-ViT, and SWAG, respectively.
These experiments demonstrate the robustness of the proposed method to the pre-training methods, datasets, and backbone architectures. 

Notably, performance improvements of \method{} are remarkable with large-scale pre-trained models, such as CLIP, CLIP-ViT, and SWAG. This is consistent with our observation in Section~\ref{subsection:mi_with_oracle}.
Our method helps large-scale pre-trained models (in terms of the pre-training dataset size) not to be biased to the training source domains compared to naive fine-tuning. Especially, naive fine-tuning of CLIP-ViT (61.1\%) shows worse out-of-domain performance than fine-tuning ImageNet pre-trained model (64.2\%). 
In contrast, \method{} can leverage the pre-trained knowledge from CLIP-ViT, resulting in superior performance (73.7\%) compared with the ImageNet pre-trained model (65.9\%).
In our later analysis, we show that the knowledge of large-scale pre-trained models is more beneficial to domain generalization than the knowledge of ImageNet pre-trained models.

\begin{table}[t]
\centering
\small
\caption{\small {\bf Comparison with methods exploiting pre-trained models.} Out-of-domain accuracies on five domain generalization benchmarks are shown. 
Average accuracies and standard errors are reported from three trials.
}
\label{table:learning_from_pretrained}
\renewcommand{\arraystretch}{1.1}
\setlength{\tabcolsep}{5pt}
\setlength{\abovetopsep}{0.5em}
\begin{tabular}{l|ccccc|c} 
\toprule
\textbf{Algorithm}  & \data{PACS} & \data{VLCS} & \data{OfficeHome} & \data{TerraInc} & \data{DomainNet} & Avg.  \\
\midrule
CRD \cite{tian2019crd}             & 82.3\scriptsize{$\pm1.0$}  &  76.6\scriptsize{$\pm0.9$}  &  67.6\scriptsize{$\pm0.4$}  &  44.0\scriptsize{$\pm1.9$}  &  42.1\scriptsize{$\pm0.1$}  &  62.5  \\
VID \cite{ahn2019vid}             & 84.9\scriptsize{$\pm0.3$}  &  76.2\scriptsize{$\pm0.2$}  &  64.6\scriptsize{$\pm0.5$}  &  48.3\scriptsize{$\pm1.3$}  &  42.5\scriptsize{$\pm0.1$}  &  63.3  \\
LP-FT \cite{kumar2022iclr_finetuning_distort}             & 84.6\scriptsize{$\pm0.8$}  &  76.7\scriptsize{$\pm1.5$}  &  65.0\scriptsize{$\pm0.2$}  &  47.1\scriptsize{$\pm0.7$}  &  43.0\scriptsize{$\pm0.1$}  &  63.3  \\
L$^2$-SP \cite{xuhong2018l2sp}         & 83.6\scriptsize{$\pm0.3$}  &  78.8\scriptsize{$\pm0.4$}  &  65.0\scriptsize{$\pm0.3$}  &  47.9\scriptsize{$\pm2.1$}  &  42.5\scriptsize{$\pm0.2$}  &  63.6  \\
DELTA \cite{li2019delta}            & 83.1\scriptsize{$\pm1.1$}  &  77.7\scriptsize{$\pm0.4$}  &  68.5\scriptsize{$\pm0.3$}  &  45.7\scriptsize{$\pm0.9$}  &  42.8\scriptsize{$\pm0.1$}  &  63.6  \\
LwF \cite{li2017lwf}            & 83.1\scriptsize{$\pm0.8$}  &  77.2\scriptsize{$\pm0.7$}  &  70.0\scriptsize{$\pm0.2$}  &  49.2\scriptsize{$\pm1.2$}  &  42.7\scriptsize{$\pm0.1$}  &  64.5  \\
\textbf{\method{}}      & \textbf{85.4\scriptsize{$\pm0.4$}} & \textbf{79.0\scriptsize{$\pm0.0$}} & \textbf{70.5\scriptsize{$\pm0.4$}} & \textbf{50.4\scriptsize{$\pm1.1$}} & \textbf{44.3\scriptsize{$\pm0.2$}} & \textbf{65.9}  \\
\bottomrule
\end{tabular}
\end{table}
\subsubsection{Comparison with methods exploiting pre-trained models.}
\label{subsubsection:learning_from_pretrained_exp}
Other DG methods simply employ pre-trained models as weight initialization, while \method{} additionally exploits it in the training process. This is the first approach to exploit pre-trained models in domain generalization, but there are several studies in other fields for different purposes. Table \ref{table:learning_from_pretrained} provides a comparison of the methods applicable to our DG settings. We exclude the methods that require additional information other than pre-trained models (\eg, pre-training datasets) or are restricted to a specific model.
As shown in the table, \method{} outperforms the comparison methods with large margins. These results demonstrate the effectiveness of our method design for the out-of-domain generalization.

\subsection{Analysis of \method{}}

\begin{figure}
    \centering
    \includegraphics[width=\columnwidth]{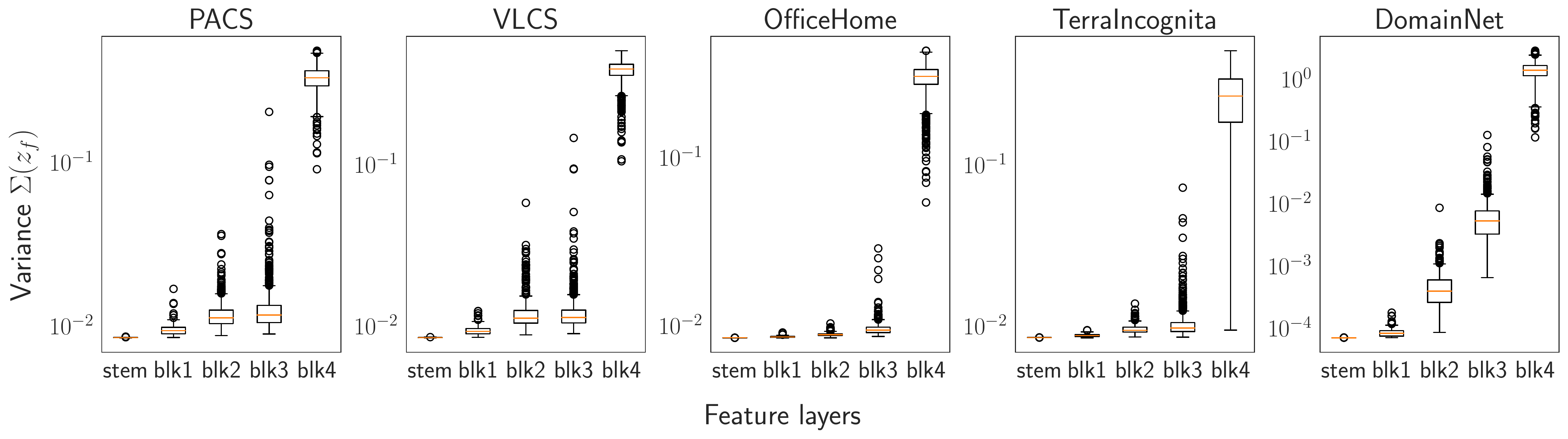}
    \caption{\small \textbf{Distribution of $\Sigma(z_f)$.} We plot the estimated variances, $\Sigma(z_f)$, for each layer. X-axis indicates the feature layer where the features $z_f$ are collected. In all datasets, the variances increase as the layer is closer to the output.}
    \label{fig:sigma_dist}
\end{figure}
\subsubsection{Loss function interpretation: $\Sigma$ distribution analysis.}
We can interpret the variance term of \method{}, $\Sigma(z_f)$ in Equation \eqref{eq:method}, as control variables of the distance loss between pre-trained features $z_{f^0}$ and current learning features $z_f$. During the training phase, if the variance values become smaller then the model will preserve MI with the pre-trained model.
On the contrary, when the model needs to learn new information, the variance will increase. We illustrate the learned variances in Figure~\ref{fig:sigma_dist}. The figure shows that pre-trained information is preserved well in lower layers, while task-specific new information is learned in higher layers.
This result is consistent with the interpretation that high layer features represent more task-specific semantic information than low layer features \cite{erhan2009visualizing}; task shifts during fine-tuning make higher layer features learn more semantics than lower layers.

\begin{table}[t]
\centering
\small
\caption{\small {\bf Performance improvements in \data{Camelyon17} medical dataset.} Even in the large distribution shift setup between pre-training and target datasets, \method{} consistently outperforms ERM.
Every accuracy is averaged over three trials.}
\label{table:camelyon}
\renewcommand{\arraystretch}{1.1}
\setlength{\tabcolsep}{5pt}
\setlength{\abovetopsep}{0.5em}
\begin{tabular}{ll|ccccc|c} 
\toprule
\textbf{Pretrain}             & \textbf{Algorithm} & 1 & 2 & 3 & 4 & 5 & Avg.  \\
\midrule\multirow{2}{*}{ImageNet ERM}   & ERM                & 97.1      & 94.7      & 95.7      & 96.4      & 90.7      & 94.9  \\
                                        & \method            & {97.5}      & 94.5      & 95.6      & {96.7}      & {93.7}      & {95.6} (+0.7)  \\
\midrule\multirow{2}{*}{SWAG}           & ERM                & 97.0      & 94.1      & 95.3      & 96.0      & 89.5      & 94.4  \\
                                        & \method            & {97.4}      & {95.5}      & {96.5}      & {96.1}      & {90.9}      & {95.3} (+0.9)  \\
\bottomrule
\end{tabular}
\end{table}
\subsubsection{Case study on \data{Camelyon17}: large distribution shift between pre-training and fine-tuning.} As shown in Equation~\eqref{eq:distribution_change}, the tightness of the lower bound is directly connected to the divergence between the representations of oracle and pre-trained models. Therefore, we investigate the case that there is a large shift between pre-trained and target datasets using the medical dataset \cite{bandi2018camelyon17,koh2021wilds}, \data{Camelyon17}. 
This dataset consists of whole-slide images of histological lymph node sections from the five hospitals, where each hospital corresponds to each domain.
The task is to predict whether the image contains tumor tissue of breast cancer. There is a large gap between the pre-training distribution (ImageNet or Instagram-3.6B) and the fine-tuning distribution (\data{Camelyon17}). Detailed visual examples are provided in Appendix \ref{subsec:imagenet_vs_camelyon17}. The results in Table~\ref{table:camelyon} demonstrate \method{} leads the model to learn robust representations even in the large distribution shift setup between pre-training and fine-tuning.

\begin{figure}
    \centering
        \includegraphics[width=\columnwidth]{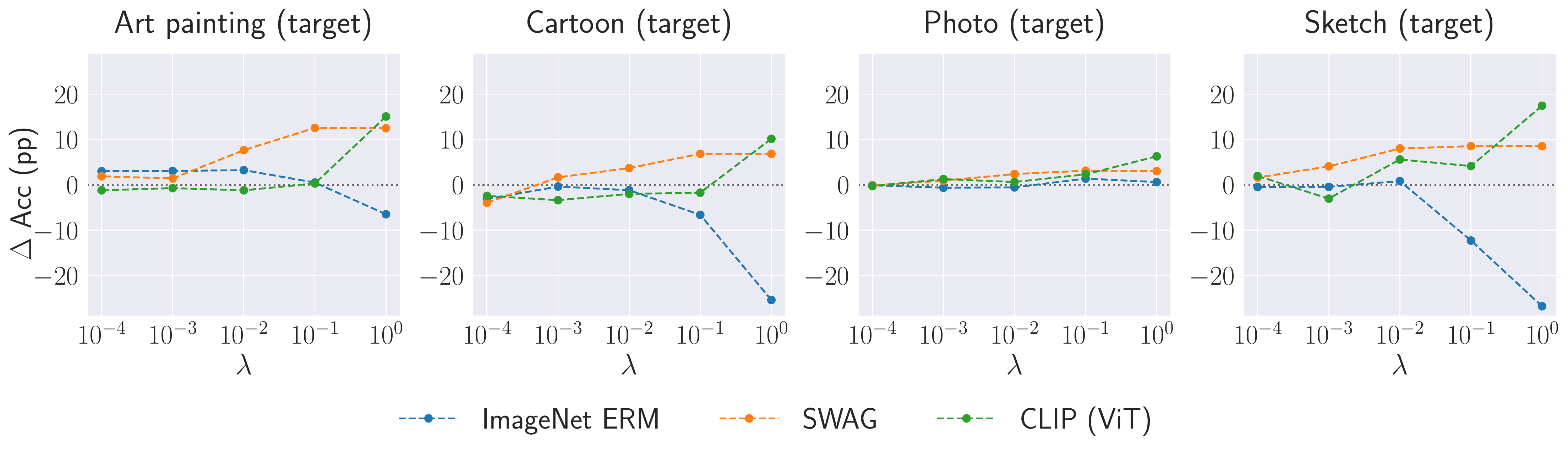}
    \caption{\small \textbf{Comparison of three pre-trained models according to $\lambda$.} Y-axis indicates the performance difference of \method{} to ERM. $\lambda$ is the intensity of the mutual information regularization. We compare three models: ResNet-50 pre-trained in ImageNet \cite{he2016_cvpr_resnet}, RegNetY-16GF pre-trained by SWAG \cite{singh2022swag}, and ViT-B pre-trained by CLIP \cite{radford2021clip}.}
    \label{fig:accdiff_varying_lambda}
\end{figure}
\subsubsection{Relationship between the pre-training scale and the intensity of the MI regularization.}
Our method has a control parameter $\lambda$, which controls the balance between the cross-entropy loss and the MI regularization loss. If $\lambda$ becomes larger, it implies that the strength of MI regularization becomes stronger, while it weakens the strength of the ERM objective. Intuitively, if the pre-trained knowledge is informative enough to the target task, larger $\lambda$ will improve the performances, while if the pre-trained knowledge is uninformative to the target task, then larger $\lambda$ can harm the performances, because of the penalty on the ERM objective.
We compare three pre-trained models (ImageNet pre-trained model, SWAG, and CLIP-ViT) by varying $\lambda$.
Figure~\ref{fig:accdiff_varying_lambda} shows how the out-of-domain performance of \method{} with different pre-trained backbones changes by $\lambda$. The additional results on different datasets are given in Appendix \ref{subsec:accdiff_by_lambda_more}.

First, we observe that the ImageNet pre-trained backbone has a negative correlation between the performance difference and $\lambda$ in target domains. When distribution shifts significantly differ, such as cartoon and sketch domains, we can observe an apparent negative correlation. We presume that it is because the ImageNet samples barely contain non-photo images, such as art painting or sketch images. On the other hand, we observe that \method{} with SWAG and CLIP-ViT backbones make significant performance improvements by choosing larger $\lambda$. In other words, SWAG and CLIP-ViT pre-trained knowledge are helpful to learn robust features for various target domains compared to the ImageNet pre-trained model. Furthermore, it implies that larger pre-trained models trained with massive diverse domain images show less sensitivity to the choice of $\lambda$, not only bringing remarkable performance improvements as shown in Table \ref{table:varying_dataset_method_backbone_scale}.

\section{Conclusion}
Traditional domain generalization (DG) approaches focus to learn a robust representation using multiple source domains.
However, in the recent trends of scaling up pre-training, the use of a large-scale pre-trained model becomes more important than the use of DG algorithms for the real-world DG.
In line with this trend, we propose \fullname{} (\method{}) to robustly exploit the pre-trained model by approximating an oracle model.
To do this, we first re-formulate the domain generalization objective by introducing a concept of an oracle model.
Then, we derive a tractable variational bound of the objective by approximating the oracle model with the pre-trained model.
Our experimental results demonstrate both the effectiveness and the potential of the proposed method. \method{} achieves state-of-the-art performance in the DomainBed benchmarks. Furthermore, when combining \method{} with large-scale pre-trained backbones, such as CLIP \cite{radford2021clip} or SWAG \cite{singh2022swag}, the performance improvements remarkably increases. 
We hope that this study promotes a new research direction of exploiting pre-trained backbones to learn robust representations for domain generalization.

\section*{Acknowledgement}
This work was supported by IITP grant funded by the Korea government (MSIT) (No. 2021-0-01341, AI Graduate School Program, CAU).

\clearpage

%
%
\bibliographystyle{splncs04}
\bibliography{egbib}

\clearpage
\newtheorem{assumption}{Assumption}

\appendix

\section{Derivation of Lower Bound}
\label{sec:derivation_of_lower_bound}
\begin{assumption}
The variational distribution $q(\cdot|z)$ satisfies the regularity condition such that, 
for any $\mathbb{P}_{X|z}\in\{\mathbb{P}'_{X|z}\mid \mathbb{E}_{X|z}[|X|^{2}] < \infty\}$,
\begin{align}
    \mathbb{E}_{X|z}\left[\left(\nabla_{x} \log q(x|z)|_{x=X}\right)^{\intercal}\nabla_{x} \log q(x|z)|_{x=X}\right] < \infty,
\end{align}
where $\mathbb{E}_{X|z}$ is a conditional expectation of $X$ given $z$.
\end{assumption}
\begin{remark}
Note that the Gaussian distribution used in our implementation satisfies the regularity condition.
To check the regularity condition of Gaussian distribution, we first compute the gradient as follows,
\begin{align}
    \nabla_{x}& \log q(x|z)|_{x=X}\\
    &= \nabla_{x} \left(C+\frac{1}{2}\log |\Sigma(z)| + \frac{1}{2}(x-\mu(z))^{\intercal}\Sigma(z)^{-1}(x-\mu(z))\right) |_{x=X}\\
    &= \Sigma(z)^{-1}(X-\mu(z)).
\end{align}
Hence, we get,
\begin{align}
    \mathbb{E}_{X|z}&\left[\left(\nabla_{x} \log q(x|z)|_{x=X}\right)^{\intercal}\nabla_{x} \log q(x|z)|_{x=X}\right]\\
    &= \mathbb{E}_{X|z}\left[(X-\mu(z))^{\intercal}\Sigma(z)^{-2}(X-\mu(z))\right] < \infty.
\end{align}
since $\mu(z)$ and $\Sigma(z)$ are finite and $\mathbb{E}_{X|z}[|X|^{2}]$ is bounded.
Hence, the Gaussian distribution satisfies the regularity condition.
\end{remark}

Under the assumption of $q$, we derive the lower bound.
\begin{proof}[Derivation of the Lower Bound]
Based on the regularity condition, we derive the lower bound of the term, $\mathbb{E}_{Z_{f^{*}},Z_{f}}\left[\log q(Z_{f^{*}}\mid Z_{f})\right]$.
Before starting the derivation, let us define $d_{2,\infty}(f,g):=\sup_{x}\|f(x)-g(x)\|_{2}$.
Then, the derivation starts from Taylor's theorem for a differentiable multivariate function.
From Taylor's theorem, there exists a point $c$ such that $c = tx + (1-t)x_{0}$ for some $t\in[0,1]$ and the following equality holds,
\begin{align}
    \log q(x\mid y) = \log q(x_{0}\mid y) + \nabla_{x} \log q(x\mid y)|_{x=c}^{\intercal}(x-x_{0}).
\end{align}
Then, we can derive the following upper bound as follows,
\begin{align}
    \log q(x\mid y) &= \log q(x_{0}\mid y) + \nabla_{x} \log q(x\mid y)|_{x=c}^{\intercal}(x-x_{0})\\
    &\leq \log q(x_{0}\mid y) + \left|\nabla_{x} \log q(x\mid y)|_{x=c}^{\intercal}(x-x_{0})\right|\\
    &\leq \log q(x_{0}\mid y) + \left\|\nabla_{x} \log q(x\mid y)|_{x=c}\right\|_{2}\left\|x-x_{0}\right\|_{2}
\end{align}
By using this bound, we can derive the following lower bound,
\begin{align}
\mathbb{E}_{Z_{f^{*}}, Z_f}&[\log q(Z_{f^{*}}\mid Z_f)]=\mathbb{E}_{X, X'}\left[\log q(f^{*}(X)\mid f(X'))\right]\\
\geq& \mathbb{E}_{X, X'}\left[\log q(f^{0}(X)\mid f(X'))\right] \nonumber\\
&- \mathbb{E}_{X, X'}\left[\left\|\nabla \log q(c(X)\mid f(X'))\right\|_{2}\left\|f^{0}(X)-f^{*}(X)\right\|_{2}\right]\\
\geq& \mathbb{E}_{X, X'}\left[\log q(f^{0}(X)\mid f(X'))\right] \nonumber\\
&- \mathbb{E}_{X, X'}\left[\left\|\nabla \log q(c(X)\mid f(X'))\right\|_{2}\right]d_{2,\infty}(f^{*},f^{0})\\
\geq& \mathbb{E}_{Z_{f^{0}}, Z_f}\left[\log q(Z_{f^{0}}\mid Z_f)\right]- Cd_{2,\infty}(f^{*},f^{0}),
\end{align}
where $c(x)$ is the function between $f^{0}$ and $f^{*}$, which selects the point satisfying Taylor's theorem, and $C$ is a constant derived from the regularity condition.
\end{proof}

\section{Additional Implementation Details}
\label{sec:additional_impl_details}

\subsection{Hyperparameter tuning}


We split the hyperparameters (HPs) into two groups: algorithm-specific HPs and algorithm-agnostic HPs. The algorithm-agnostic HPs consist of batch size, learning rate, dropout, and weight decay, and \method{} has only one algorithm-specific HP, $\lambda$.
To reduce the computational cost, we tune the algorithm-specific HPs and algorithm-agnostic HPs independently. 
We first search algorithm-specific HPs with default algorithm-agnostic HPs, then search algorithm-agnostic HPs with the tuned algorithm-specific HPs.
That is, the $\lambda$ is searched in [1.0, 0.1, 0.01, 0.001] with the batch size of 32, the learning rate of 5e-5, no dropout, and no weight decay. Then, we search algorithm-agnostic HPs with the searched $\lambda$ following Cha \etal \cite{cha2021swad}. They propose reduced HP search space for efficiency compared to DomainBed \cite{gulrajani2020domainbed}. The protocol searches the learning rate in [1e-5, 3e-5, 5e-5], dropout in [0.0, 0.1, 0.5], and weight decay in [1e-4, 1e-6]. The batch size per domain is fixed to 32. Since \method{} is a regularization method, we add a case of no weight decay.

Even though we use the efficient HP search protocol, it still requires heavy computational resources. Therefore, we tune $\lambda$ only for the non-main experiments, including combination with SWAD, combination with various pre-trained backbones, and the case study on \data{Camelyon17}. Also, we use the batch size of 16 for SWAG \cite{singh2022swag} due to the GPU memory limitation. Note that there is room for further performance improvement by intensive HP tuning and additional usage of GPU memory, considering the simplified HP search protocol and limited computational resources.

\subsection{Implementation details}

The variance encoder is initialized to estimate the variance of 0.1. It is chosen by observing the convergence point of the variance.
Softplus function is employed to ensure non-negativity of the variance. Also, we empirically apply the 10 times larger learning rate for the mean and variance encoders than the feature extractor and the classifier.

\subsection{Mutual information estimation}
\label{subsec:mine}

In Section \ref{subsection:mi_with_oracle}, we estimate the mutual information using Mutual Information Neural Estimator (MINE) \cite{belghazi2018mine}. The mutual information is estimated by MINE as follows:

\begin{align}
    \widehat{I(Z_{f^*} ; Z_f})=\sup _{\theta \in \Theta} \mathbb{E}_{\mathbb{P}_{Z_{f^*} Z_f}}\left[T_{\theta}\right]-\log \left(\mathbb{E}_{\mathbb{P}_{Z_{f^*}} \otimes {\mathbb{P}}_{Z_f}}\left[e^{T_{\theta}}\right]\right).
\end{align}


For the features $Z_{f^*}$ and $Z_f$, the features after global average pooling are uniformly collected by domains. The statistics network, $T_\theta$, consists of two hidden linear layers with 512 dimensions and ELU activation functions, following \cite{belghazi2018mine}. In the case of fine-tuning, such as ERM$-$, ERM$+$, and \method{}, the models are trained as many as the number of target domains. Therefore, we estimate the mutual information for each model and report their average value.

\section{Additional Analysis and Discussion}

\subsection{Variations on the assumptions of domain generalization}
\begin{table}[h]
\centering
\small
\caption{\small {\bf Performances of class-conditional MIRO.} 
}
\label{table:cmiro}
\renewcommand{\arraystretch}{1.2}
\setlength{\tabcolsep}{5pt}
\begin{tabular}{l|cccc|c} 
\toprule
\textbf{Algorithm}   & \pacs{}          & \vlcs{}          & \oh{}            & \data{TerraInc}            & \textbf{Avg.}           \\ \midrule
ERM    & 84.2\scriptsize{$\pm0.1$}          & 77.3\scriptsize{$\pm0.1$}          & 67.6\scriptsize{$\pm0.2$}          & 47.8\scriptsize{$\pm0.6$}          & 69.2           \\
MIRO   & \textbf{85.4}\scriptsize{$\pm0.4$}          & \textbf{79.0}\scriptsize{$\pm0.0$} & 70.5\scriptsize{$\pm0.4$}          & \textbf{50.4}\scriptsize{$\pm1.1$} & \textbf{71.3}  \\
C-MIRO & 85.3\scriptsize{$\pm0.5$} & 78.5\scriptsize{$\pm0.5$}          & \textbf{70.8}\scriptsize{$\pm0.3$} & 49.4\scriptsize{$\pm0.3$}          & 71.0 \\
\bottomrule
\end{tabular}
\end{table}



In general, domain generalization (DG) assumes that there are multiple source domains, source domain labels are available, and the same input has the same label between source and target domains. Here, we can make the variations on the problem settings by changing the assumption. Single-source DG does not assume the multiple source domains \cite{cha2021swad,fan2021adversarially_single_dg}. Several studies try to solve DG problem without domain labels \cite{carlucci2019jigsaw_jigen,cha2021swad,matsuura2020aaai_mixture_mld}. Heterogeneous DG deals with the label set shift, \ie, the same input can have different labels between source and target domains \cite{li2019heteroDG,wang2020interdomain_mixup_icassp_dg}. In this task, it is assumed that a classifier is learnable in the target domain and the methods focus on the feature extractor. The proposed method exploits pre-trained models instead of assuming available multiple source domains or source domain labels, and focuses on the feature extractor instead of the classifier. Therefore, \method{} is directly applicable to single-source DG, DG without domain labels, and heterogeneous DG problems.
On the other hand, we can consider a more specific type of distribution shift. In this case, we may need a different mutual information (MI) strategy. For example, we can employ class-conditional MI for class-conditional distribution shift (C-MIRO) by using $I(Z_{f^*};Z_f|Y)$ instead of $I(Z_{f^*}; Z_f)$.
In Table \ref{table:cmiro}, C-MIRO achieves comparable scores with MIRO and outperforms ERM even though the problem setting is not class-conditional. From the results, we believe that MIRO can be adapted to other distribution shifts by choosing the proper MI strategy.

\subsection{The relationship between mutual information and domain generalization performance}
\label{subsec:relationship_mi_dg}

\begin{table}
\centering
\centering
\small
\caption{\small {\bf Average accuracies of ERM$-$, ERM$+$, and MIRO in \pacs{}.} ERM$-$ and ERM$+$ indicate ERM without and with pre-trained model, respectively.}
\label{table:mi_acc_corr}
\renewcommand{\arraystretch}{1.1}
\setlength{\tabcolsep}{4pt}
\begin{tabular}{l|ccc} 
\toprule
\textbf{Pre-trained model}             & \textbf{ERM$-$} & \textbf{ERM$+$} & \textbf{MIRO}  \\ \midrule
ResNet-50 \scriptsize{(ImageNet)}         & 51.6 & 84.2 & 85.4  \\
RegNet-16GF \scriptsize{(Instagram-3.6B)} & 51.5 & 89.6 & 97.4  \\
\bottomrule
\end{tabular}
\end{table}

Our method assumes that knowledge of the oracle model helps domain generalization and it is transferable to the target model by maximizing mutual information (MI). These assumptions are quite intuitive, but there is no theoretical guarantee that MI with the oracle model is perfectly aligned with DG performance. Instead, we empirically observe that a high MI model shows better DG performance if the empirical loss constraint of Equation (2) holds; the pre-trained model itself has high MI but does not satisfy this constraint.
Figure \ref{fig:mi} in the main text shows the rankings of MI for ERM$-$, ERM$+$, and MIRO are in order. Table \ref{table:mi_acc_corr} shows that the rankings are the same for accuracies: ERM$-$ (51.6\%), ERM$+$ (84.2\%), and MIRO (85.4\%) in ImageNet pre-trained ResNet and ERM$-$ (51.5\%), ERM$+$ (89.6\%), and MIRO (97.4\%) in Instagram-3.6B pre-trained RegNet, respectively.

\section{Additional Results}

\subsection{Visual comparison between ImageNet and \data{Camelyon17}}
\label{subsec:imagenet_vs_camelyon17}

\begin{figure}
\begin{center}
    \begin{subfigure}{0.4\linewidth}
        \includegraphics[width=\linewidth]{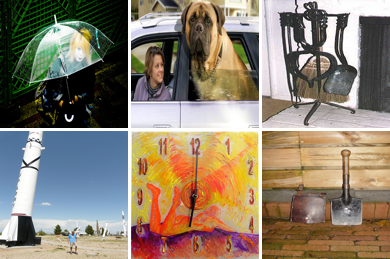}
        \caption{{ImageNet} (pre-train)}
        \label{fig:imagenet}
    \end{subfigure}
    \quad
    \begin{subfigure}{0.4\linewidth}
        \includegraphics[width=\linewidth]{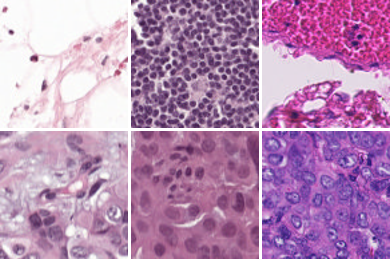}
        \caption{\data{Camelyon17} (fine-tuning)}
        \label{fig:camelyon}
    \end{subfigure}

    \caption{\small \textbf{Example images of ImageNet and \data{Camelyon17}.} Large distribution shift occurs between pre-training ({ImageNet}) and fine-tuning (\data{Camelyon17}). ImageNet is a multiclass objective recognition task and \data{Camelyon17} is a binary classification task for reading whether the image contains tumor tissue. Instagram-3.6B examples are omitted since it is not publicly available.
    }
    \label{fig:imagenet_camelyon}
\end{center}
\end{figure}

Figure \ref{fig:imagenet_camelyon} shows a huge visual gap between pre-training (ImageNet) and fine-tuning (\data{Camelyon17}) datasets. The tasks are also different; ImageNet is an object recognition task and \data{Camelyon17} is a binary classification of breast cancer. Despite the large gap between pre-training and fine-tuning distribution, the proposed method shows consistent performance improvement (See Table \ref{table:camelyon} in the main text).


\subsection{Relationship between the pre-training scale and the intensity of the mutual information regularization}
\label{subsec:accdiff_by_lambda_more}

\begin{figure}
    \centering
    \begin{subfigure}{\linewidth}
        \includegraphics[width=\columnwidth]{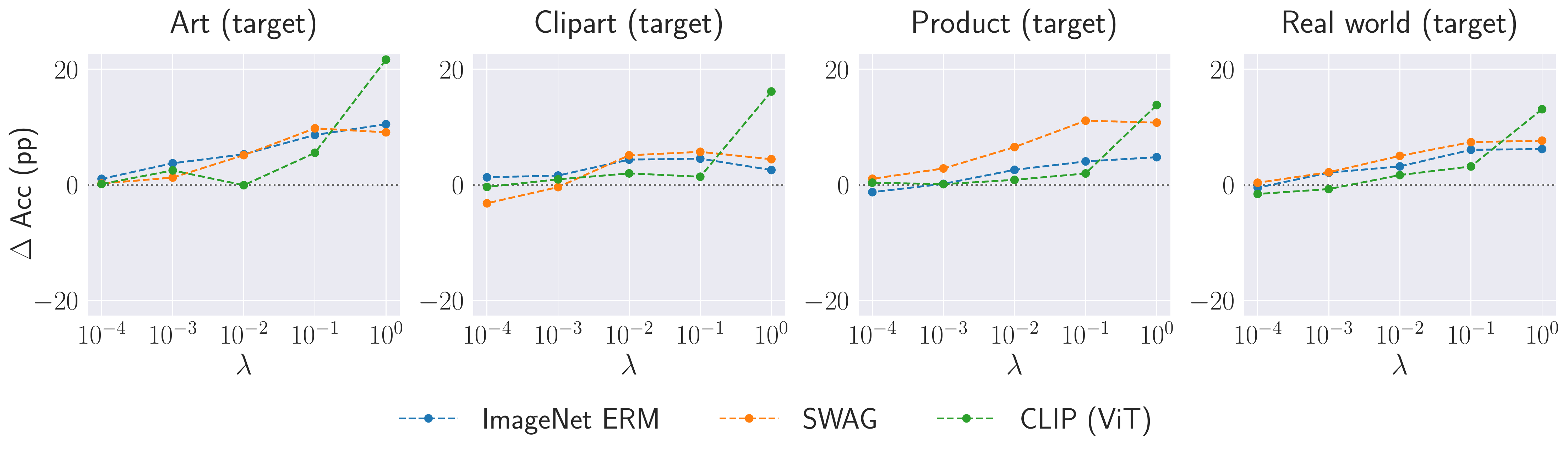}
        \caption{\small \oh}
    \end{subfigure}
    \begin{subfigure}{\linewidth}
        \includegraphics[width=\columnwidth]{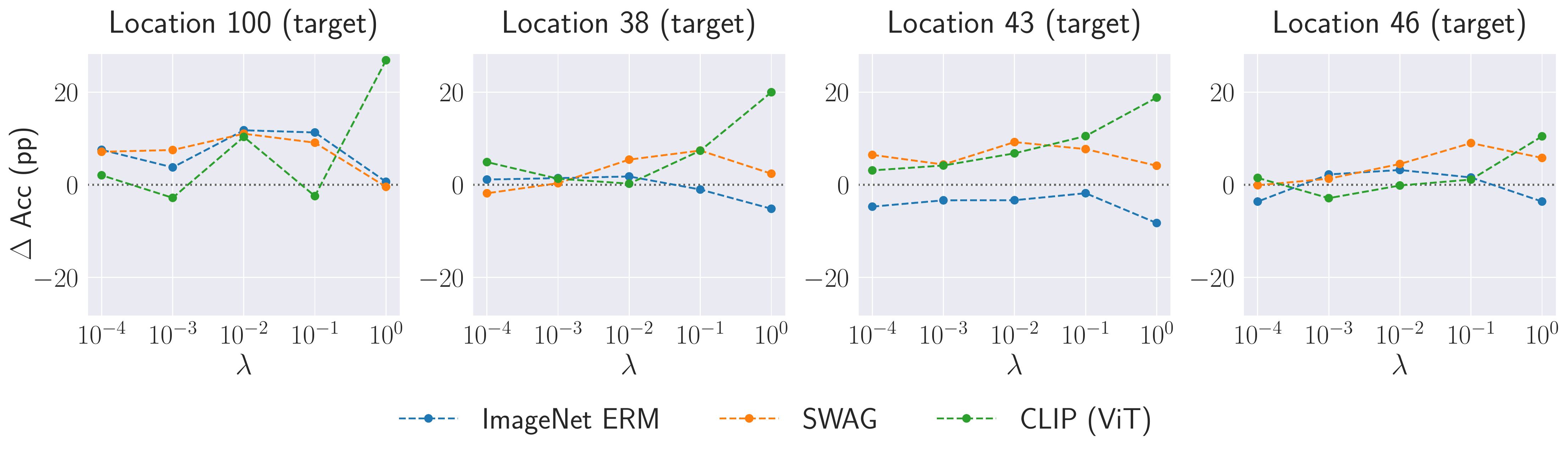}
        \caption{\small \tr}
    \end{subfigure}
    \begin{subfigure}{0.75\linewidth}
        \includegraphics[width=\columnwidth]{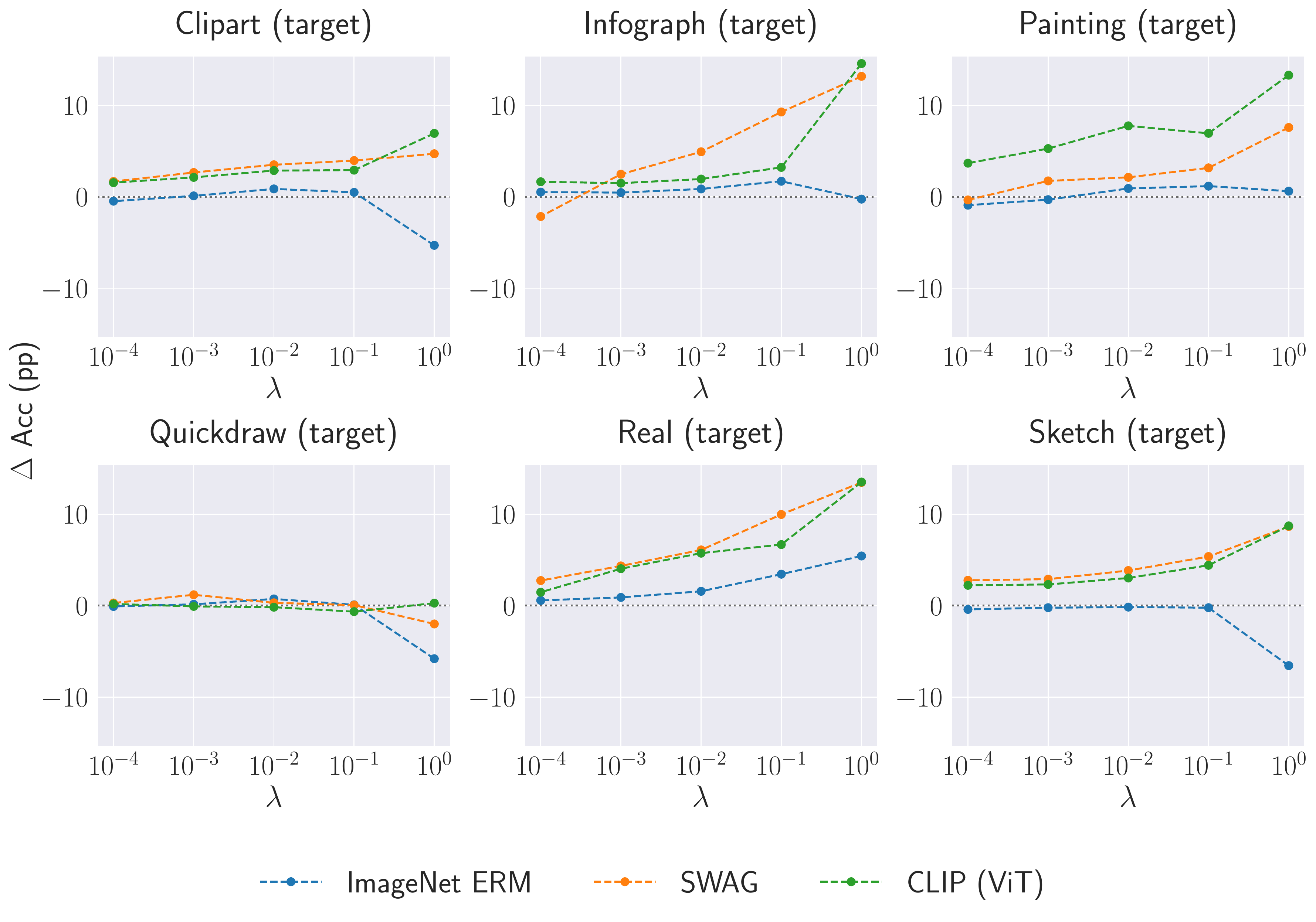}
        \caption{\small \dn}
    \end{subfigure}

    \caption{\small \textbf{Comparison of three pre-trained models according to $\lambda$.} Y-axis indicates the performance difference of \method{} to ERM. $\lambda$ is the intensity of the mutual information regularization. We compare three models: ResNet-50 pre-trained in ImageNet \cite{he2016_cvpr_resnet}, RegNetY-16GF pre-trained by SWAG \cite{singh2022swag}, and ViT-B pre-trained by CLIP \cite{radford2021clip}.}
    \label{fig:accdiff_varying_lambda_more}
\end{figure}

In this section, we provide the extended results of Figure \ref{fig:accdiff_varying_lambda} in the main text. Figure \ref{fig:accdiff_varying_lambda_more} shows the additional comparison of three pre-trained backbones according to $\lambda$ about \oh{}, \tr{}, and \dn{}. The comparisons show similar trends with the results in \pacs{}. ImageNet pre-trained backbone, such as ResNet-50 pre-trained in ImageNet \cite{he2016_cvpr_resnet}, has a negative correlation between the performance difference and $\lambda$ in some target domains. Large-scale pre-trained backbones, such as SWAG \cite{singh2022swag} and CLIP \cite{radford2021clip}, tend to consistently make significant performance improvements at high $\lambda$ and become less sensitive to the choice of $\lambda$. 



\end{document}